\documentclass[a4paper,11pt]{article}
\usepackage[utf8]{inputenc}
\usepackage[T1]{fontenc} 
\usepackage{float} 
\usepackage{multirow}
\usepackage{cite}

\makeatletter
\gdef\@fpheader{ }
\gdef\@journal{ }
\makeatother
\RequirePackage{amsmath}
\RequirePackage{amssymb}
\RequirePackage{epsfig}
\RequirePackage{graphicx}
\RequirePackage[numbers,sort&compress]{natbib}
\RequirePackage{color}
\RequirePackage[colorlinks=true
,urlcolor=blue
,anchorcolor=blue
,citecolor=blue
,filecolor=blue
,linkcolor=blue
,menucolor=blue
,pagecolor=blue
,linktocpage=true
,pdfproducer=medialab
,pdfa=true
]{hyperref}

\newif\ifnotoc\notocfalse
\newif\ifemailadd\emailaddfalse
\newif\iftoccontinuous\toccontinuousfalse
\makeatletter
\def\@subheader{\@empty}
\def\@keywords{\@empty}
\def\@abstract{\@empty}
\def\@xtum{\@empty}
\def\@dedicated{\@empty}
\def\@arxivnumber{\@empty}
\def\@collaboration{\@empty}
\def\@collaborationImg{\@empty}
\def\@proceeding{\@empty}
\def\@preprint{\@empty}

\newcommand{\subheader}[1]{\gdef\@subheader{#1}}
\newcommand{\keywords}[1]{\if!\@keywords!\gdef\@keywords{#1}\else%
\PackageWarningNoLine{\jname}{Keywords already defined.\MessageBreak Ignoring last definition.}\fi}
\renewcommand{\abstract}[1]{\gdef\@abstract{#1}}
\newcommand{\dedicated}[1]{\gdef\@dedicated{#1}}
\newcommand{\arxivnumber}[1]{\gdef\@arxivnumber{#1}}
\newcommand{\proceeding}[1]{\gdef\@proceeding{#1}}
\newcommand{\xtumfont}[1]{\textsc{#1}}
\newcommand{\correctionref}[3]{\gdef\@xtum{\xtumfont{#1} \href{#2}{#3}}}
\newcommand\jname{JHEP}

\newcommand\preprint[1]{\gdef\@preprint{\hfill #1}}

\makeatother
\newcommand\note[2][]{%
\if!#1!%
\stepcounter{footnote}\footnotetext{#2}%
\else%
{\renewcommand\thefootnote{#1}%
\footnotetext{#2}}%
\fi}

\makeatletter
\newtoks\auth@toks
\renewcommand{\author}[2][]{%
  \if!#1!%
    \auth@toks=\expandafter{\the\auth@toks#2\ }%
  \else
    \auth@toks=\expandafter{\the\auth@toks#2$^{#1}$\ }%
  \fi
}
\makeatother
\makeatletter
\newtoks\affil@toks\newif\ifaffil\affilfalse
\newcommand{\affiliation}[2][]{%
\affiltrue
  \if!#1!%
    \affil@toks=\expandafter{\the\affil@toks{\item[]#2}}%
  \else
    \affil@toks=\expandafter{\the\affil@toks{\item[$^{#1}$]#2}}%
  \fi
}
\makeatother

\makeatletter
\newtoks\email@toks\newcounter{email@counter}%
\newcommand{\emailAdd}[1]{%
\emailaddtrue%
\ifnum\theemail@counter>0\email@toks=\expandafter{\the\email@toks, \@email{#1}}%
\else\email@toks=\expandafter{\the\email@toks\@email{#1}}%
\fi\stepcounter{email@counter}}
\newcommand{\@email}[1]{\href{mailto:#1}{\tt #1}}
\makeatother

\makeatletter
\newcommand*\collaboration[1]{\gdef\@collaboration{#1}}
\newcommand*\collaborationImg[2][]{\gdef\@collaborationImg{#2}}
\makeatletter

\newcommand\afterLogoSpace{\smallskip}
\newcommand\afterSubheaderSpace{\vskip3pt plus 2pt minus 1pt}
\newcommand\afterProceedingsSpace{\vskip21pt plus0.4fil minus15pt}
\newcommand\afterTitleSpace{\vskip23pt plus0.06fil minus13pt}
\newcommand\afterRuleSpace{\vskip23pt plus0.06fil minus13pt}
\newcommand\afterCollaborationSpace{\vskip3pt plus 2pt minus 1pt}
\newcommand\afterCollaborationImgSpace{\vskip3pt plus 2pt minus 1pt}
\newcommand\afterAuthorSpace{\vskip5pt plus4pt minus4pt}
\newcommand\afterAffiliationSpace{\vskip3pt plus3pt}
\newcommand\afterEmailSpace{\vskip16pt plus9pt minus10pt\filbreak}
\newcommand\afterXtumSpace{\par\bigskip}
\newcommand\afterAbstractSpace{\vskip16pt plus9pt minus13pt}
\newcommand\afterKeywordsSpace{\vskip16pt plus9pt minus13pt}
\newcommand\afterArxivSpace{\vskip3pt plus0.01fil minus10pt}
\newcommand\afterDedicatedSpace{\vskip0pt plus0.01fil}
\newcommand\afterTocSpace{\bigskip\medskip}
\newcommand\afterTocRuleSpace{\bigskip\bigskip}

\newlength{\affiliationsSep}
\newcommand\beforetochook{\pagestyle{myplain}\pagenumbering{roman}}

\DeclareFixedFont\trfont{OT1}{phv}{b}{sc}{11}

\renewcommand\maketitle{
\pagestyle{empty}
\thispagestyle{titlepage}
\setcounter{page}{0}
\noindent{\small\scshape\@fpheader}\@preprint\par

\afterLogoSpace
\if!\@subheader!\else\noindent{\trfont{\@subheader}}\fi
\afterSubheaderSpace
\if!\@proceeding!\else\noindent{\sc\@proceeding}\fi
\afterProceedingsSpace
{\LARGE\flushleft\sffamily\bfseries\@title\par}
\afterTitleSpace
\hrule height 1.5\p@%
\afterRuleSpace
\if!\@collaboration!\else
{\Large\bfseries\sffamily\raggedright\@collaboration}\par
\afterCollaborationSpace
\fi
\if!\@collaborationImg!\else
{\normalsize\bfseries\sffamily\raggedright\@collaborationImg}\par
\afterCollaborationImgSpace
\fi
{\bfseries\raggedright\sffamily\the\auth@toks\par}
\afterAuthorSpace
\ifaffil\begin{list}{}{%
\setlength{\leftmargin}{0.28cm}%
\setlength{\labelsep}{0pt}%
\setlength{\itemsep}{\affiliationsSep}%
\setlength{\topsep}{-\parskip}}
\itshape\small%
\the\affil@toks
\end{list}\fi
\afterAffiliationSpace
\ifemailadd 
\noindent\hspace{0.28cm}\begin{minipage}[l]{.9\textwidth}
\begin{flushleft}
\textit{E-mail:} \the\email@toks
\end{flushleft}
\end{minipage}
\else 
\PackageWarningNoLine{\jname}{E-mails are missing.\MessageBreak Plese use \protect\emailAdd\space macro to provide e-mails.}
\fi
\afterEmailSpace
\if!\@xtum!\else\noindent{\@xtum}\afterXtumSpace\fi
\if!\@abstract!\else\noindent{\renewcommand\baselinestretch{.9}\textsc{Abstract:}}\ \@abstract\afterAbstractSpace\fi
\if!\@keywords!\else\noindent{\textsc{Keywords:}} \@keywords\afterKeywordsSpace\fi
\if!\@arxivnumber!\else\noindent{\textsc{ArXiv ePrint:}} \href{http://arxiv.org/abs/\@arxivnumber}{\@arxivnumber}\afterArxivSpace\fi
\if!\@dedicated!\else\vbox{\small\it\raggedleft\@dedicated}\afterDedicatedSpace\fi
\ifnotoc\else
\iftoccontinuous\else\newpage\fi
\beforetochook\hrule
\tableofcontents
\afterTocSpace
\hrule
\afterTocRuleSpace
\fi
\setcounter{footnote}{0}
\pagestyle{myplain}\pagenumbering{arabic}
} 

\renewcommand{\baselinestretch}{1.1}\normalsize
\divide\@tempdima\baselineskip
\@tempcnta=\@tempdima
\skip\@mpfootins = \skip\footins
\renewcommand{\@dotsep}{10000}

\newcommand\ps@myplain{
\pagenumbering{arabic}
\renewcommand\@oddfoot{\hfill-- \thepage\ --\hfill}
\renewcommand\@oddhead{}}
\let\ps@plain=\ps@myplain

\newcommand\ps@titlepage{\renewcommand\@oddfoot{}\renewcommand\@oddhead{}}

\numberwithin{equation}{section}

\renewcommand\section{\@startsection{section}{1}{\z@}%
                                   {-3.5ex \@plus -1.3ex \@minus -.7ex}%
                                   {2.3ex \@plus.4ex \@minus .4ex}%
                                   {\normalfont\large\bfseries}}
\renewcommand\subsection{\@startsection{subsection}{2}{\z@}%
                                   {-2.3ex\@plus -1ex \@minus -.5ex}%
                                   {1.2ex \@plus .3ex \@minus .3ex}%
                                   {\normalfont\normalsize\bfseries}}
\renewcommand\subsubsection{\@startsection{subsubsection}{3}{\z@}%
                                   {-2.3ex\@plus -1ex \@minus -.5ex}%
                                   {1ex \@plus .2ex \@minus .2ex}%
                                   {\normalfont\normalsize\bfseries}}
\renewcommand\paragraph{\@startsection{paragraph}{4}{\z@}%
                                   {1.75ex \@plus1ex \@minus.2ex}%
                                   {-1em}%
                                   {\normalfont\normalsize\bfseries}}
\renewcommand\subparagraph{\@startsection{subparagraph}{5}{\parindent}%
                                   {1.75ex \@plus1ex \@minus .2ex}%
                                   {-1em}%
                                   {\normalfont\normalsize\bfseries}}

\def\fnum@figure{\textbf{\figurename\nobreakspace\thefigure}}
\def\fnum@table{\textbf{\tablename\nobreakspace\thetable}}

\long\def\@makecaption#1#2{%
  \vskip\abovecaptionskip
  \sbox\@tempboxa{\small #1. #2}%
  \ifdim \wd\@tempboxa >\hsize
    \small #1. #2\par
  \else
    \global \@minipagefalse
    \hb@xt@\hsize{\hfil\box\@tempboxa\hfil}%
  \fi
  \vskip\belowcaptionskip}

\renewenvironment{thebibliography}[1]{%
\begin{oldthebibliography}{#1}%
\small%
\raggedright%
\setlength{\itemsep}{5pt plus 0.2ex minus 0.05ex}%
}%
{%
\end{oldthebibliography}%
}

\begin{document}


\title{\boldmath An Unsupervised Deep-Learning Method for Bone Age Assessment}

\author[a]{Hao Zhu,}
\author[a,1]{Wan-Jing Nie,}\note{Wan-Jing Nie and Hao Zhu contribute equivalently.}
\author[a]{Yue-Jie Hou,}
\author[a]{Qi-Meng Du,}
\author[a,2]{Si-Jing Li,}\note{lisijing@dali.edu.cn}
\author[a,3]{and Chi-Chun Zhou}\note{zhouchichun@dali.edu.cn}

\affiliation[a]{School of Engineering, Dali University, Dali, Yunnan 671003, PR China}








\abstract{The bone age, reflecting the degree of development of the bones, 
can be used to predict 
the adult height and detect endocrine diseases of children. 
Both examinations of radiologists and variability of operators have a significant impact on bone age assessment. To decrease human intervention , machine learning algorithms are used to assess the 
bone age automatically.
However, conventional supervised deep-learning methods need pre-labeled data.
In this paper,  
based on the convolutional auto-encoder with constraints (CCAE), an unsupervised deep-learning model proposed in the classification of the fingerprint, we 
propose this model for the  classification of the bone age and baptize it BA-CCAE. In the proposed BA-CCAE model, the key regions of the raw X-ray images of the bone age are encoded, yielding the latent vectors. 
The K-means clustering algorithm is used to obtain the final classifications by grouping the latent vectors of the bone images.  
A set of experiments on the Radiological 
Society of North America pediatric bone age dataset  (RSNA) show that  
the accuracy of classifications at $48$-month intervals 
is $76.15\%$. Although the accuracy now is lower than most of the existing supervised 
models, the proposed BA-CCAE model can establish the classification
of bone age without any pre-labeled data, and to the best of our 
knowledge, the proposed BA-CCAE is one of the few trails using the 
unsupervised deep-learning method for the bone age assessment.
}
\keywords{bone age assessment, unsupervised learning, clustering algorithm}

\maketitle
\flushbottom


\section{Introduction}
Bone age is a scale for assessing developmental maturity and detecting the extent of 
skeletal development. It offers vital information in many areas, such as selecting  
artistic talents \cite{mao2011application} and the forensic determination of the 
age of a specific individual as a basis for a conviction \cite{raja2010pyrolysis}.  
Usually, the X-rays of the left hand can better reflect the growth, 
development level, and maturity of an individual \cite{gilsanz2005hand}. 

There are many conventional methods including 
the counting method \cite{vogt1938osseous}, the Greulich and Pyle (GP) atlas method \cite{anderson1971use}, 
the Tanner-Whitehouse (TW)  method \cite{malina2002assessment}, and etc. 
The counting method evaluates bone age by counting the numbers of the ossification center \cite{vogt1938osseous}. 
The GP atlas method compares the X-ray Images of the bone with the standard bone to obtain the bone age \cite{anderson1971use}. 
The TW method is based on the comparison between the specific target bone and the score table to obtain the bone age \cite{mughal2014bone}. 
The conventional methods vary from each other due to the differences in races and regions \cite{ontell1996bone}.
For example, in China, the Chinese Carpal Bone Development Standard-CHN Method method \cite{zhang2008standards} is developed based on the TW method, where the main changes are removing the ulna bones and adding the cephalic and uncinate bones.

Beyond the conventional method, supervised machine learning methods, 
including deep-learning method, 
are used in the bone age assessment.
In those approaches, pre-labeled X-ray Images are required for the training set.
They are used to train the algorithms and then 
the trained models are utilized to gain the bone ages from X-ray Images of the test set.
In the classification tasks, the model gives an interval of the bone age, say from $24$th 
to $48$th months, and the accuracy (ACC) is used to evaluate the performance of the model.
In the regression task, the mean absolute error (MAE) is used.
For example, three supervised models, C19-SVM, C19-Res, and C19-ST-Res, are used to categorize the bone age.
The experiments on $12,536$  DICOM data (the training set contrains $10,656$ images and the validation set
contains $1,880$ images) collected by the orthopedic picture archive and communication system (PACS) of Shengjing Hospital of China Medical University show that the
top1  matching accuracies are $55.2\%$, $60.9\%$, and $62.6\%$
\cite{chen2020automatic} respectively.
A deep convolutional neural network (CNN) based on fine-grained image classification for automatic bone age assessment is proposed in Ref. \cite{li2021automatic} and achieves an accuracy of $66.38\%$ for males and $68.63\%$ for females, and the MAE are $3.71 \pm 7.55$ and $3.81 \pm 7.74$ months for males and females, respectively.
By using the convolutional neural network (CNN), Ref. \cite{reddy2020bone} reports
the MAE of the bone age assessment $4.7$ months with the whole hand, and $5.1$ months with the index finger.
In $2020$, Chen Chao et al. propose an attention-guided approach to automatically localize the discriminative regions for bone age assessment
achieving the MAE $4.3$ months  \cite{chen2021attention}. 
In $2021$, Zhang Y et al. extracted skeletal features based on the inception V$3$ neural network, 
and use the adversarial regression learning network (ARLNet) to obtain an MAE of $3.01$ months \cite{zhang2021adversarial}.
Currently, there is a system for bone age assessment based on the 
CHN method achieving an accuracy of $93\%$ on 
the Chinese children data set \cite{yin2022computerized}.

In this paper, instead of the supervised deep-learning method, we proposed 
an unsupervised deep-learning method for the bone age assessment.
The main method is based on the previous work \cite{hou2021unsupervised},
where a convolutional auto-encoder with constraints (CCAE), 
an unsupervised deep-learning model, is proposed in the classification of the fingerprint.
The propose unsupervised deep-learning model for the classification of the bone age is 
named BA-CCAE for short. In this approach, each key regions of the raw X-ray images of the bone age are encoded by a specified CCAE, yielding a collection of the latent vectors. And then the K-means clustering algorithm is used to obtain the final classifications by grouping the latent vectors of the bone images.  
A set of experiments on the Radiological 
Society of North America pediatric bone age dataset  (RSNA) show that  
the accuracy of classifications at $48$-month intervals 
is $76.15\%$. It suggests that the BA-CCAE is more accurate than CCAE in bone age classification.
Although the accuracy now is lower than most of the existing supervised 
models, the proposed BA-CCAE model can establish the classification
of bone age without any pre-labeled data.

\section{The Dataset and Data Preprocessing}
In this section, we give an introduction of the dataset and the data preprocessing.
The data preprocessing is important for the proposed method.

\subsection{The Dataset: the Data Selection and the Classification }
\begin{figure}
 \centering
\includegraphics[width=0.9\textwidth]{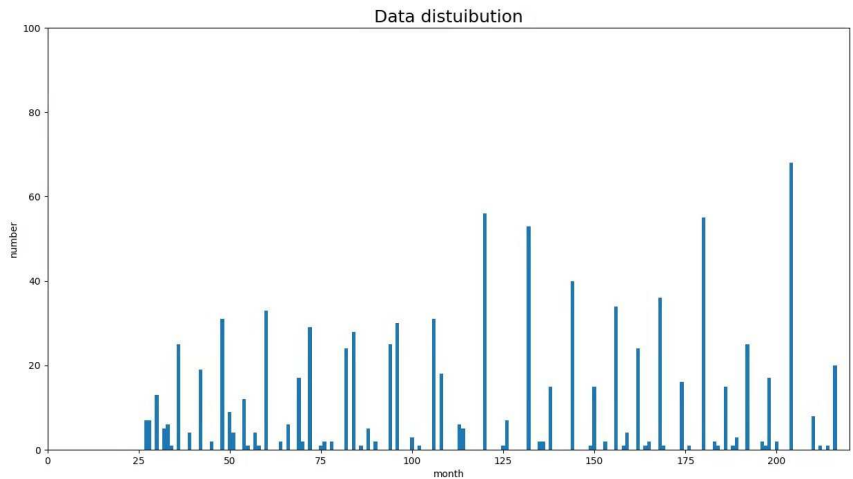}
\caption{The distribution of the bone age of selected $960$ dataset.}
\label{label1}
\end{figure}

Given that in the existing supervised machine learning methods of bone age assessment, 
the test sets are usually consisted of about $1,000$ subsamples. 
For example, the test sets in Refs. \cite{reddy2020bone,chen2021attention,zhang2021adversarial,yin2022computerized}
are with sizes $307$, $200$, $500$, and $200$ respectively.
In this work, without loss of generality, we use only a portion of the 
bone age data selected from the RSNA in order to show the effectiveness of the proposed 
unsupervised deep-learning method.
As a result,
$960$ X-ray Images of the bone ranging from $24$ months to $216$ months are selected.
Here, the task of bone age assessment is converted into a classification task.
We take $48$ months as an interval yielding a dataset with $4$ labels. For example, 
the first classes consisting of bones with bone age whithn $24$th to $72$th months.
An overview of the selected data is given in Fig. (\ref{label1}) where the distribution
of the bone age is given.
The data is available at 
\href{https://www.rsna.org/education/ai-resources-and-training/ai-image-challenge/RSNA-Pediatric-Bone-Age-Challenge-2017}{RSNA Pediatric Bone Age Challenge (2017)}. 

\subsection{The Data Preprocessing: the Clipping of Key Regions and Other Treatments}
\begin{figure}
 \centering
\includegraphics[width=0.4\textwidth]{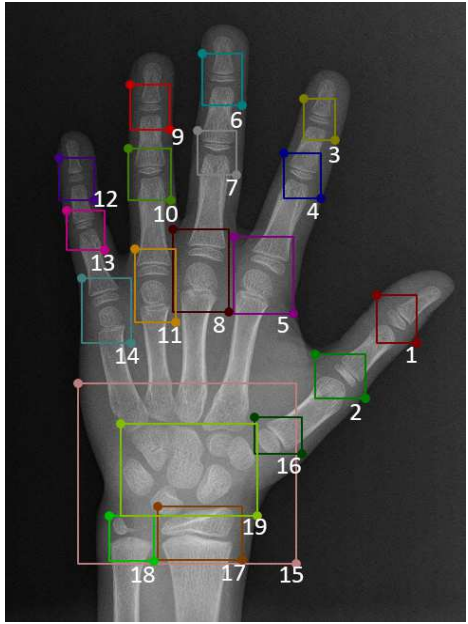}
\caption{An example of the segmentation of the key regions of the X-ray Images of the bones.}
\label{label}
\end{figure}

The main ideal of the proposed BA-CCAE is to encode each key regions
by an independent CCAE.
In this section, we manually crop out the key regions  
of the X-ray Images.
In the following section, selected key regions 
are encoded by the CCAE, in order to establish the classification automatically.
As shown in Fig. (\ref{label}),
the key regions are joints and are labeled by index from $1$ to $19$.
For example, the joint of thumb is labeled by $1$ and the carpal bone 
position without background is labeled by  $19$. 
The clipping will be replaced by machine learning method in the following research.

\begin{figure}
 \centering
\includegraphics[width=1.0\textwidth]{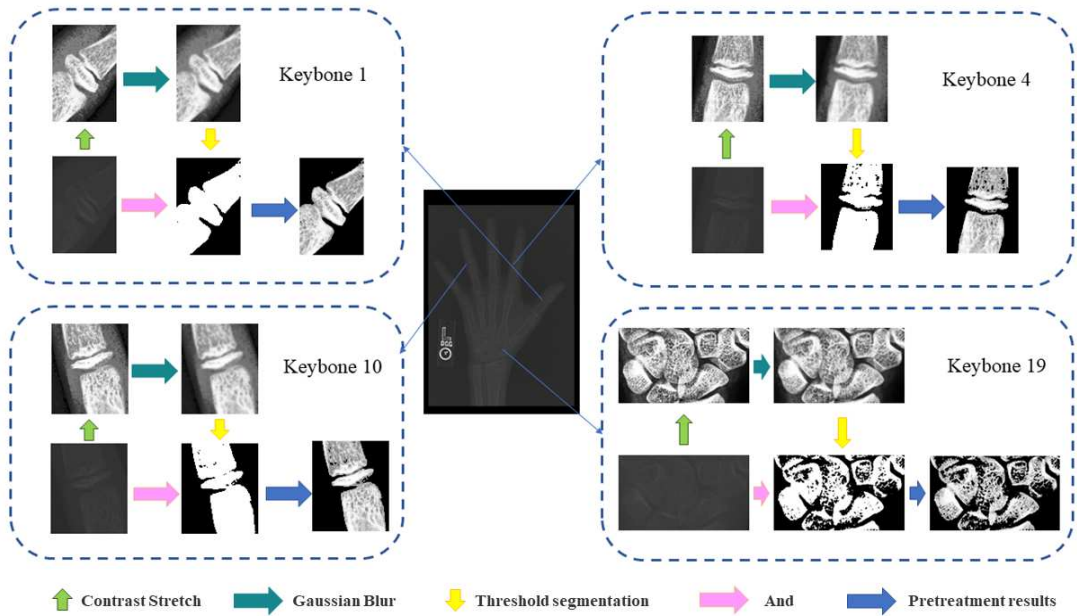}
\caption{An flowchart of the data preprocessing.}
\label{yuchuli}
\end{figure}

Then, the preprocess of contrast equalization, Gaussian blurring, and adaptive threshold segmentation 
are applied, in order to obtain data with high quality.
In the contrast equalization, the histogram equalization \cite{pizer1987adaptive} is applied where 
an nonlinear transformation
is used to transform the distribution 
of pixels into an uniform distribution.
The purpose of Gaussian blurring is to reduce the noise \cite{li2017gaussian}.
Finally, adaptive threshold segmentation transforms the X-ray Images
into a binarized image, in order to separate the bones from the background. 
It shows in Fig. (\ref{yuchuli}) that the interference from background 
and annotation information is effectively eliminated.

\section{The Main Method}
\begin{figure}
\centering
\includegraphics[width=0.85\textwidth]{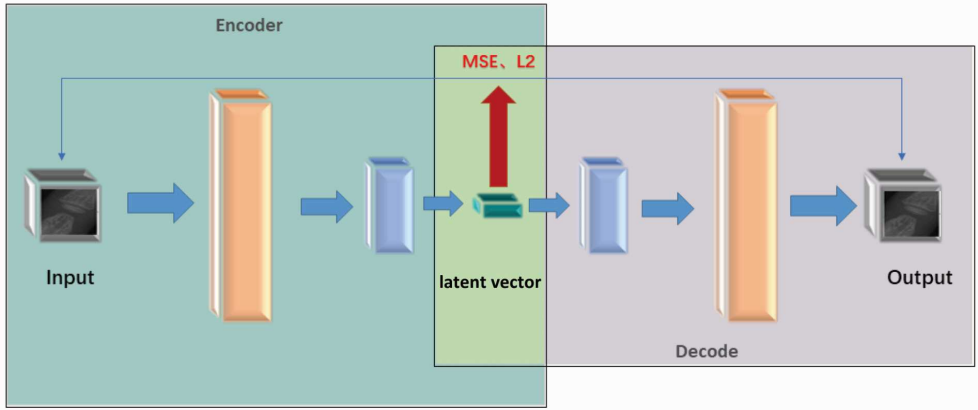}
\caption{A diagram of the basic structure of the CCAE.}
\label{CCAE}
\end{figure}

In this section, we introduce the proposed BA-CCAE model.
It is based on the CCAE proposed in the previous work \cite{hou2021unsupervised}.
The CCAE is an effective model that can extract key features from a given images
and compress the dimension of the raw images by giving the encoded vectors, as shown in 
 Fig. (\ref{CCAE}).
In the CCAE, an L2 constraint of the latent vector is added to the 
mean square error function between the decoded and raw images giving the final loss function.  
More details of the CCAE can be found in the previous work \cite{hou2021unsupervised}.

\begin{figure}
\centering
\includegraphics[width=0.9\textwidth]{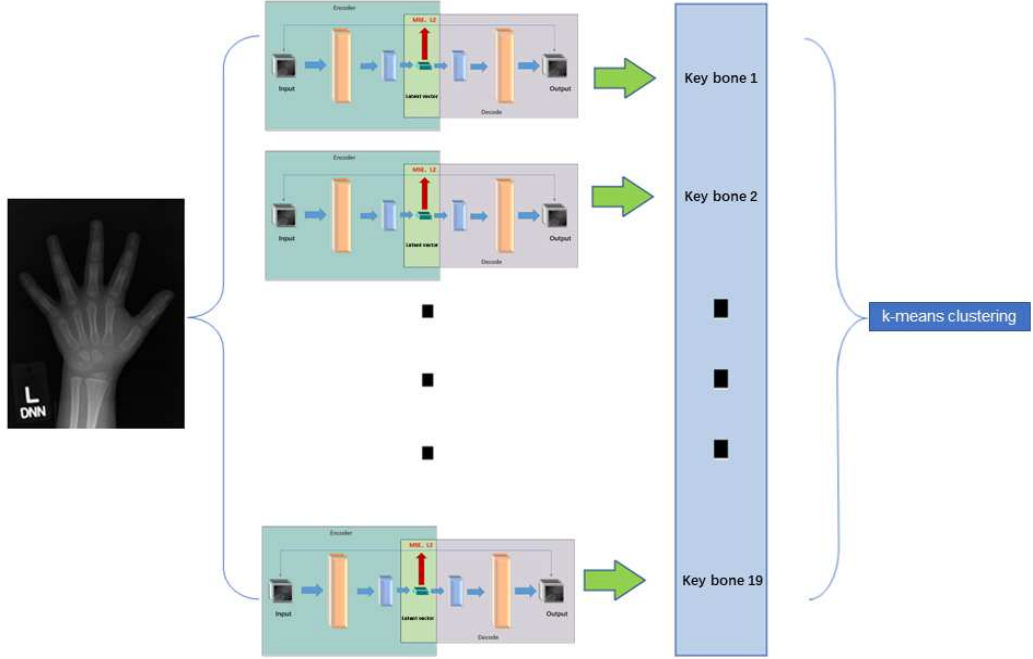}
\caption{A diagram of the BA-CCAE.}
\label{BACCAE}
\end{figure}

The main ideal of the BA-CCAE is to learn the key features of key regions independently
by using CCAE with different hyper parameters to encode selected images of key regions.
In the BA-CCAE model, each CCAEs have the convolutional layers with 
kernel size $30\times30$. Due to the different image sizes, such as 
images of the knuckles and carpal bones, 
the hyper parameters of each CCAEs are differently chosen 
to fit the X-ray Images of the bone.
As shown in Fig. (\ref{BACCAE}),
selected key regions are encoded by independent CCAEs resulting in a 
collection of encoded latent vectors.

Finally, the K-means algorithm \cite{macqueen1967some} 
is used to obtain the final cluster, or classification, based on the encoded latent vectors.

\section{Results and Analyses}
In this section, the results and the analyses of the BA-CCAE is given.

\subsection{Results: the Accuracies}
By using the combination of key regions with index $2$, $5$, $8$, $11$, $16$, 
$17$, and $19$, 
the K-means gives best classification accuracy at interval $48$ months.
In this approach, the K-means are used to group the $960$ samples into $16$ 
groups. In each group, we count the distribution of real labels and obtain the 
the dominated label. The dominated label is chosen to be the label given by the proposed 
method.
As a result, an overall accuracy of $76.15\%$ on the  $960$ samples taking $48$ months as interval is obtained.
The detail of the clustering result is given in Tab. (\ref{table1611}) 

\begin{table}[htbp]
\centering
\caption{The detail of the clustering result. 
The combination of key regions with index $2$, $5$, $8$, $11$, $16$, 
$17$, and $19$, the interval is $48$ months.}
\begin{tabular}{ccccccc}
\hline
Group & $24$th to  & $72-$ & $120-$ & $168-$ & Dominated & Accuracy  \\
labels &$71$th months &119  &167  &216  & label &   \\
\hline
$0$ & $0$ & $2$ & $8$ &  $\mathbf{ 25}$ & $168-216$ &$71.43\%$ \\
$1$ & $0$ & $4$ & $\mathbf{50}$ & $13$ & $120-168$ &$74.62\%$\\
$2$ & $9$ & $\mathbf{63}$ & $9$ & $0$ & $72-120$ &$77.78\%$\\
$3$ & $0$ & $3$ & $10$ & $\mathbf{48}$ & $168-216$ &$78.69\%$\\
$4$ & $2$ & $\mathbf{35}$ & $32$ & $0$ & $72-120$& $50.72\%$\\
$5$ & $\mathbf{38}$ & $5$ & $1$ & $0$ & $24-72$&$86.36\%$ \\
$6$ & $0$ & $0$ & $28$ & $\mathbf{38}$ & $168-216$&$57.58\%$ \\
$7$ & $0$ & $10$ & $\mathbf{62}$ & $7$ & $120-168$ &$78.48\%$\\
$8$ & $0$ & $0$ & $10$ & $\mathbf{52}$ & $168-216$ &$83.87\%$\\
$\mathbf{9} $& $4$ & $2$ & $1$ & $0$ & $24-72$&$57.14\%$\\
$10$ & $\mathbf{59}$ & $6$ & $0$ & $0$ & $24-72$&$90.77\%$\\
$11$ & $0$ & $0$ & $7$ & $\mathbf{90} $& $168-216$ &$92.78\%$\\
$12$ & $\mathbf{55}$& $15$ & $0$ & $0$ & $24-72$ &$78.57\%$\\
$13$ & $1$ & $10$ & $\mathbf{35}$ & $3$ & $120-168$ &$71.42\%$\\
$14$ & $\mathbf{21}$ & $2$ & $0$ & $0$ & $24-72$ &$91.30\%$\\
$15$ & $22$ & $\mathbf{56}$ & $7$ & $0$ & $72-119$ &$65.88\%$\\
\hline
&   &   &  & & Overall accuracy &  $76.15\%$\\
\hline
\end{tabular}
\label{table1611}
\end{table} 

\begin{figure}
\centering
 \setlength{\abovecaptionskip}{0.cm}
 \setlength{\belowcaptionskip}{-0.cm}
\includegraphics[width=0.85\textwidth]{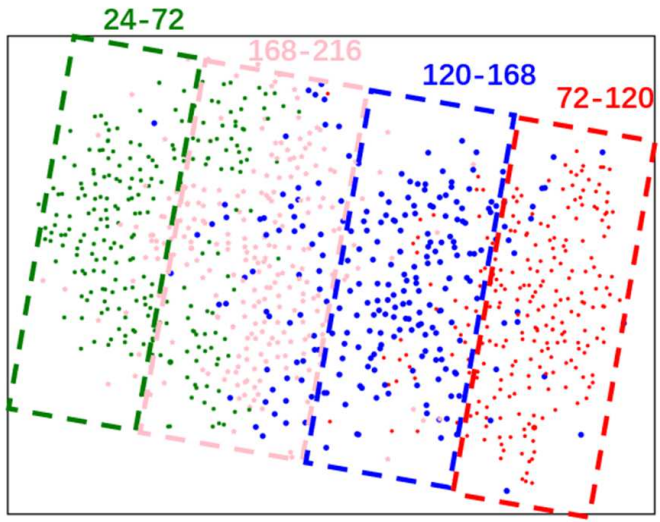}
\caption{t-SNE map of bone age results}
\label{tsne}
\end{figure}

Here, we use a t-SNE visualization graph \cite{hinton2002stochastic}, an effective method that maps the
high dimensional data into a two-dimensional plain and can provide a clear view of the
clustering result,
to show the result of the 
BA-CCAE, as shown in Fig. (\ref{tsne}). 

\subsection{ Analyses: the Contribution of Different Key Region}
\begin{figure}
\centering
\includegraphics[width=1.0\textwidth]{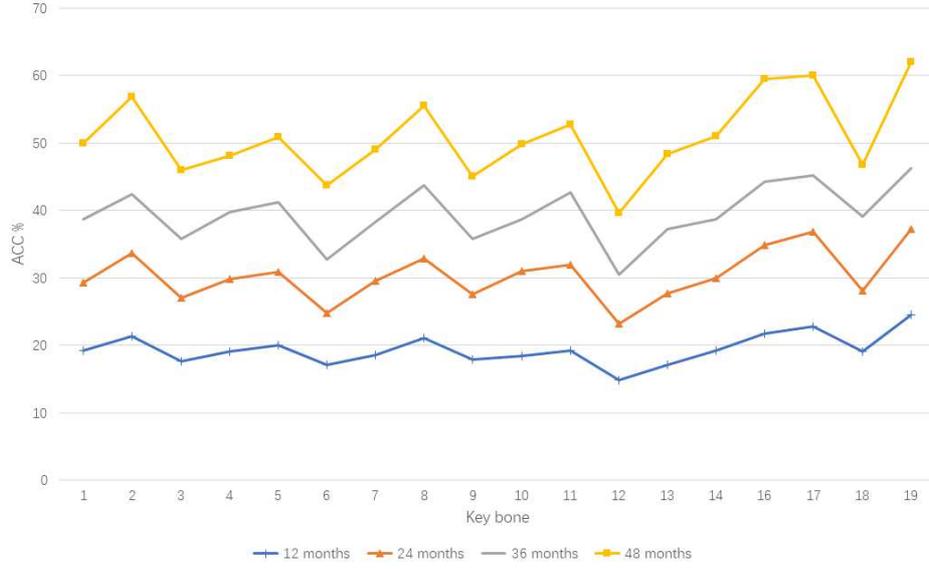}
\caption{Accuracies evaluated by different intervals 
of BA-CCAE model by using different single key regions as input. The number of 
clustering groups is $16$.}
\label{summary}
\end{figure}

We manually crop out $19$ key regions from the raw X-ray Images.
And it shows that 
the best result, the accuracy is $76.15\%$, is the combination of key regions indexed by 
$2$, $5$, $8$, $11$, $16$, $17$, and $19$.
In this section, the contribution of each key region is evaluated by showing the 
accuracy of result of using only one key region.
The result is shown in Fig. (\ref{summary}), where an line chart showing the 
accuracies using different key regions and different intervals.
The highest accuracy of is $65.42\%$ of key region indexed by $19$ and 
the positions of key regions indexed by $2$, $5$, $8$, $11$, $17$, $19$ give local peaks. 
This explains the combination of key regions indexed by 
$2$, $5$, $8$, $11$, $16$, $17$, and $19$ is the optimal choice of the key regions.

\subsection{Analyses: the Different Combinations of Key Regions}
In this section, we give the result of the BA-CCAE by using different combination of the key regions.
Among those combinations, some are chosen according to the literature. For example,
the key region indexed by $17$ in the left hand X-ray Images 
is proved to be effective \cite{liu2018bone}, and the key regions from 
a single finger is also proved to be effective \cite{reddy2020bone}. 
The results are shown in Tab. (\ref{4}).

\begin{table}
\centering
\caption{Unsupervised accuracy is compared with the supervised MAE}\label{tab:tablenotes}

\begin{tabular}{cccc}
\hline
 dataset number & Key bone  & month & Accuracy\\ 
\hline
$960$ number & $2$, $5$, $8$, $11$, $17$, $19$ & $48$ month & $72.08\%$ \\
$960$ number & $2$, $5$, $8$, $11$, $16$, $19$ & $48$ month & $73.33\%$ \\
$960$ number & $2$, $5$, $8$, $11$, $16$, $17$, $19$ & $48$ month & $76.15\%$ \\
$960$ number & $2$, $5$, $6$, $7$, $8$, $11$, $14$, $17$, $18$, $19$ \cite{liu2018bone} & $48$ month & $69.69$\% \\
$960$ number & $3$, $4$ \cite{reddy2020bone} & $48$ month & $49.48\%$ \\
\hline
\end{tabular} 
\label{4}
\end{table}

\subsection{Analyses: the Influence of Target Group Numbers of the K-means}
In this section, we give the result of the BA-CCAE with different target 
group numbers in the K-means algorithm.
The results are shown in Tab. (\ref{5}) where the target group numbers 
of $4$, $8$, $16$, $24$, $32$, and $64$ are considered. The ACC and the MAE 
are given to evaluate the influence of target group numbers.
It shows that the larger the number of clusters, the higher accuracy rate. 
However, to cluster the bones into more groups will not help
the doctors in bone age assessment. Because the doctor needs to check more groups of data.
Therefore, we chose the target group numbers $16$ .

\begin{table}[!ht]
\caption{U-net autoencoder clustering accuracy results of bone age.}
\label{tab:tablenotes}
\centering 
\begin{tabular}{ccccccc}
\hline
 Number of cluster classes & $4$-group & $8$-group & $16$-group &  $24$-group & $32$-group & $64$-group\\ 
\hline
$48$ month ACC & $64.27\%$ & $67.81\%$ & $76.15\%$ & $74.06\%$ & $74.90\%$ & $76.88\%$ \\
$48$ month MAE & $98.25$ & $67.18$ & $35.7$ & $28$ & $28.22$ & $17.38 $\\
\hline
\end{tabular}
\label{5}
\end{table} 
  
\subsection{Analyses: Comparison between the Variational Auto-Encoder (VAE) and the BA-CCAE } 
The CCAE is an variation of auto-encoder (AE). The VAE is another variation of the AE and 
is an important generative  model \cite{kingma2013auto}.
In this section, we give a comparison between the variational auto-encoder (VAE) and the BA-CCAE.
The VAE model 
The results are shown in Tab. (\ref{6}). It shows that the BA-CCAE out performs the VAE.
\begin{table}[!ht]
\caption{VAE and BA-CCAE auto encoder clustering accuracy results for bone age}\label{tab:tablenotes}
\centering

\begin{tabular}{cccccc}
\hline
Key bone & VAE ACC & BA-CCAE ACC & Key bone & VAE ACC & BA-CCAE ACC\\ 
\hline
$1$ & $47.81\%$ & $51.35$\% & $2$ & $52.19\%$ & $58.13\%$ \\
$3$ & $42.71\%$ & $47.71\%$ & $4$ & $46.56\%$ & $49.58\%$ \\
$5$ & $51.04\%$ & $52.3\%$ & $6$ & $48.44\%$ & $46.04\%$ \\
$7$ & $46.25\%$ & $50.63\%$ & $8$ & $48.96\%$ & $57.50\%$ \\
$9$ & $40.52\%$ & $45.83\%$ & $10$ & $47.29\%$ & $51.15\%$ \\
$11$ & $50.10\%$ & $54.06\%$ & $12$ & $40.63\%$ & $41.04\%$ \\
$13$ & $43.65\%$ & $49.58\%$ & $14$ & $47.81\%$ & $52.29\%$ \\
$16$ & $58.02\%$ & $65\%$ & $17$ & $53.65\%$ & $61.77\%$ \\
$18$ & $44.90\%$ & $47.92\%$ & $19$ & $60.73\%$ & $65.42\%$ \\
\hline
\end{tabular}   
\label{6}
\end{table} 

\subsection{The Result of the BA-CCAE on Dataset with Different Sizes}
\begin{figure}
\centering
\includegraphics[width=1.0\textwidth]{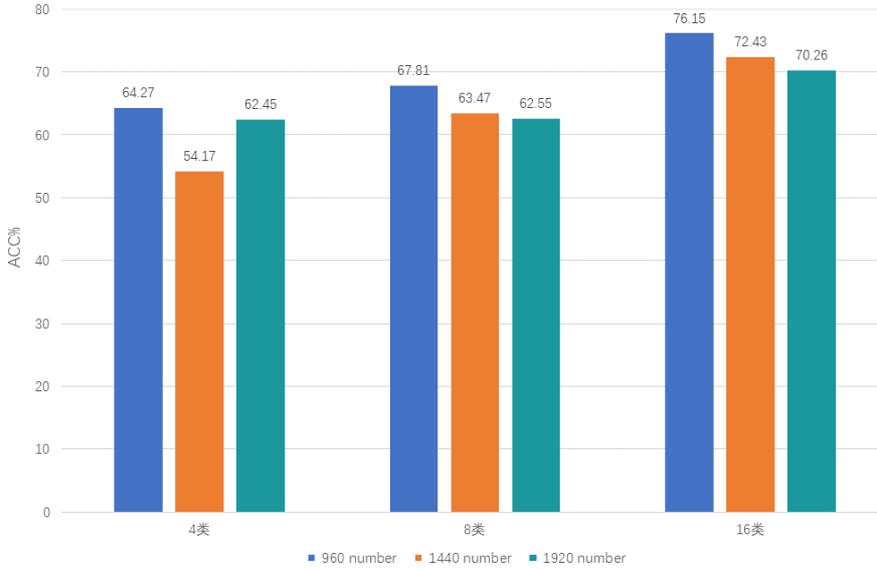}
\caption{The result of the BA-CCAE on dataset with different sizes.}
\label{6.}
\end{figure}

In this section, we give the result of the BA-CCAE on dataset with sizes $1,440$ and $1,920$, in
order to verify the robustness of the model.
It shows in Fig. (\ref{6.}) that all the accuracies increase with the target group numbers.

\begin{figure}
\centering
\includegraphics[width=1.0\textwidth]{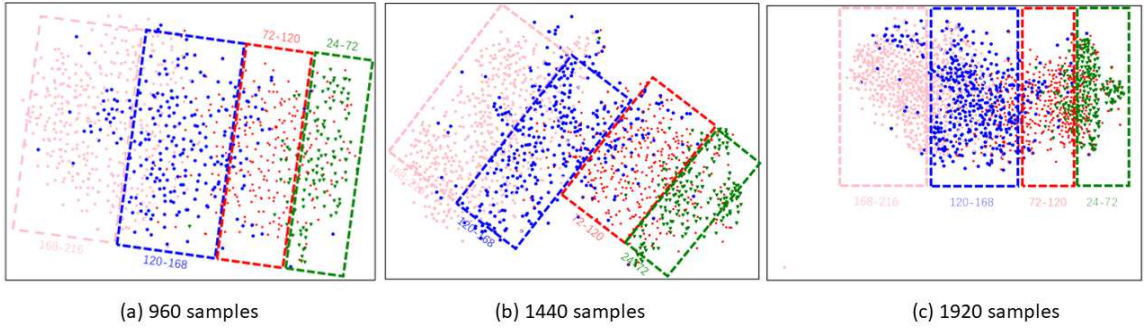}
\caption{The t-SNE visualization graphs with data sizes $960$, $1,440$, and $1,920$ respectively.}
\label{10.}
\end{figure}

\begin{figure}
\centering
\includegraphics[width=0.9\textwidth]{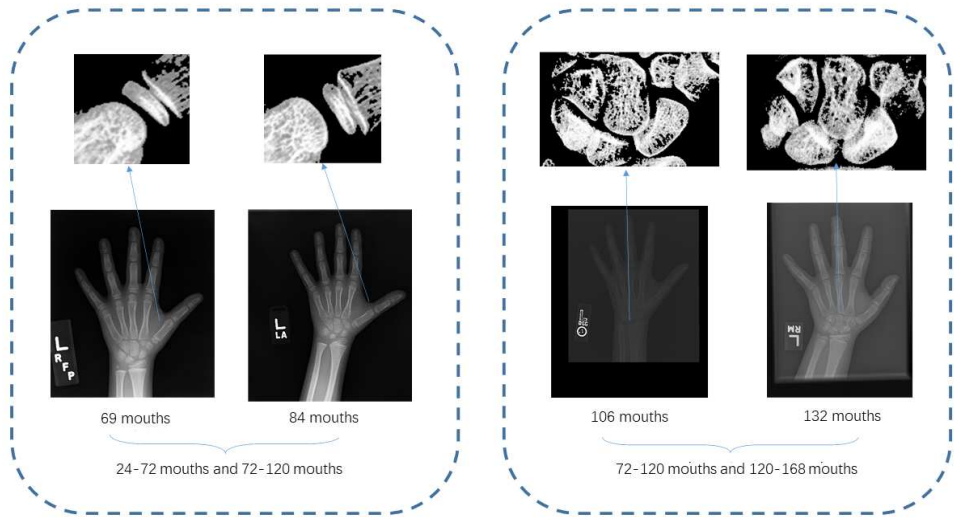}
\caption{Examples of bone that are mis-labeled  by the BA-CCAE. For those X-ray Images,
it is difficult to distinguish between them even by visual inspection.}
\label{13.}
\end{figure}

However at target group numbers $16$, the accuracies decrease with the data sizes.
To show the reason, the t-SNE visualization graphs with different data sizes are given in Fig. (\ref{10.}). It shows that as the data sizes increase, the boundary between different classes 
becomes blurred. By searching and analyzing the original bone age data that fail to cluster,
it is found that the bone age data onto different months has similar key bone features, 
as shown in Fig. (\ref{13.}). The results show that the some bone age images are  highly similar, 
and it is difficult to distinguish it by manual observation. 

\section{Conclusion and Discussions}
In this paper, instead of the supervised machine learning method, we propose an 
unsupervised deep-learning method, the BA-CCAE model, for the bone age assessment.
The result on $960$ samples selected from the RSNA dataset show that
the combination of the key regions indexed by $2$, $5$, $8$, $11$, $16$, $17$, and $19$
gives the accuracy of $76.15\%$ with the target cluster number $16$.

Although the proposed unsupervised method gives classification accuracy
lower than those existing supervised method,
it needs no pre-labeled data as the training set.
To the best of our knowledge, this work is  one of the few trials
using of the unsupervised method
for the bone age assessment.
In the following research, we will development the unsupervised method.

\section{Acknowledgments}
We are very indebted to Prof. Wu-Sheng Dai for his enlightenment and encouragement. 
We are very indebted to Profs. Guan-Wen Fang and Yong Xie for his encouragement. 
This work is supported by National Natural Science Funds of China (Grant No. 62106033), Yunnan Youth
Basic Research Projects (202001AU070020), and Doctoral Programs of Dali University
(KYBS201910).











\end{document}